\documentclass[pmlr]{jmlr}

\usepackage{longtable}

\usepackage{booktabs}
\usepackage[load-configurations=version-1]{siunitx} 

\makeatletter
\def\set@curr@file#1{\def\@curr@file{#1}} 
\makeatother

\theorembodyfont{\upshape}
\theoremheaderfont{\scshape}
\theorempostheader{:}
\theoremsep{\newline}

\jmlrvolume{}
\jmlryear{2020}
\jmlrworkshop{Machine Learning for Healthcare}

\usepackage{multirow}
\usepackage{array}
\usepackage{subcaption}
\usepackage{soul}
\usepackage{float}
\usepackage{graphicx}
\title[Minimally-trained Blood Pressure Estimation]{Developing Personalized Models of Blood Pressure Estimation from Wearable Sensors Data Using Minimally-trained Domain Adversarial Neural Networks}

\author{\Name{Lida Zhang$^1$}
      \Email{lidazhang@tamu.edu}
       \AND
       \Name{Nathan C. Hurley}$^1$
       \Email{natech@tamu.edu}
       \AND
        \Name{Bassem Ibrahim$^2$}
      \Email{bassem@tamu.edu}
      \AND
       \Name{Erica Spatz$^3$}
      \Email{erica.spatz@yale.edu}
      \AND
      \Name{Harlan M. Krumholz$^3$}
      \Email{harlan.krumholz@yale.edu}
       \AND
             \Name{Roozbeh Jafari$^{4,1,2}$}
      \Email{rjafari@tamu.edu}
      \AND
       \Name{Bobak J. Mortazavi}$^{1}$
       \Email{bobakm@tamu.edu}
       \AND
       \addr $^1$Department of Computer Science and Engineering, Texas A\&M University, USA\\
          \addr $^2$Department of Electrical and Computer Engineering, Texas A\&M University, USA\\
      \addr $^3$Yale School of Medicine, Yale University, USA\\
      \addr $^4$Department of Biomedical Engineering, Texas A\&M University, USA\\}

\begin{document}

\maketitle

\begin{abstract}
Blood pressure monitoring is an essential component of hypertension management and in the prediction of associated comorbidities.  Blood pressure is a dynamic vital sign with frequent changes throughout a given day.  Capturing blood pressure remotely and frequently (also known as ambulatory blood pressure monitoring) has traditionally been achieved by measuring blood pressure at discrete intervals using an inflatable cuff.
However, there is growing interest in developing a cuffless ambulatory blood pressure monitoring system to measure blood pressure continuously. One such approach is by utilizing bioimpedance sensors to build regression models. A practical problem with this approach is that the amount of data required to confidently train such a regression model can be prohibitive. In this paper, we propose the application of the domain-adversarial training neural network (DANN) method on our multitask learning (MTL) blood pressure estimation model, allowing for knowledge transfer between subjects.
Our proposed model obtains average root mean square error (RMSE) of $4.80 \pm 0.74$ mmHg for diastolic blood pressure and $7.34 \pm 1.88$ mmHg for systolic blood pressure when using three minutes of training data, $4.64 \pm 0.60$ mmHg and $7.10 \pm 1.79$ respectively when using four minutes of training data, and $4.48 \pm 0.57$ mmHg and $6.79 \pm 1.70$ respectively when using five minutes of training data.
DANN improves training with minimal data
in comparison to both directly training and to training with a pretrained model from another subject, decreasing RMSE by $0.19$ to $0.26$ mmHg (diastolic) and by $0.46$ to $0.67$ mmHg (systolic) in comparison to the best baseline models.
We observe that four minutes of training data is the minimum requirement for our framework to exceed ISO standards within this cohort of patients.
\end{abstract}

\section{Introduction}
\label{sec:intro}

Hypertension is a worldwide chronic disease that causes an estimated 7.6 million deaths every year. The diagnosis of hypertension is usually based on clinical blood pressure readings, but the measurement of blood pressure outside of a clinical visit (also known as ambulatory blood pressure measurement) can provide better prognostic guidance than measurements during a routine clinic visit \citep{ref2}, due to well-known confounders such as masked hypertension \citep{ref8}, white coat hypertension \citep{ref9}, and nocturnal non-dipping hypertension \citep{ref10}. Ambulatory blood pressure monitoring has been shown to be more predictive of cardiovascular mortality than clinical monitoring in a study of 63,910 adults \citep{ref17}, and nocturnal measurements are likely stronger predictors of cardiovascular risk than diurnal monitoring \citep{ref18,ref19,ref20}. Therefore, increased ambulatory measuring is desirable for public health. However, on-market ambulatory monitoring devices are not appropriate for extensive use for a number of reasons: they require specific patient postures, they are obtrusive, they disrupt sleep, and they result in poor adherence. Cuffless blood pressure monitoring devices are desirable for their possibility to overcome each of those shortcomings.
Cuffless blood pressure estimation techniques utilize devices to monitor surrogates of blood pressure, and use these surrogates to build regression models to estimate diastolic and systolic blood pressure. 

There are a variety of techniques that have recently been investigated for their potential application to cuffless blood pressure estimation. Chief among those techniques include photoplethysmography (PPG) in conjunction with electrocardiography (ECG) \citep{ref26}, dual PPGs \citep{ref42, ref43}, Doppler radar technology \citep{ref27}, or bioimpedance \citep{ref28}. Each of these techniques attempt to measure the pulse transit time (PTT) or pulse wave velocity (PWV), both of which are known surrogates for blood pressure \citep{ref30,ref31,ref32}. \citet{ref31} developed a bioimpedance-based sensor that locates arterial sites to measure these physiologic surrogates of blood pressure. \citeauthor{ref31} then used a window-based AdaBoost regression technique to measure personal diastolic and systolic blood pressure over windows of 10 consecutive beats to with respective errors of 2.6 mmHg and 3.4 mmHg. This finding falls within the ISO standard requiring errors less than 10 mmHg when comparing with a gold standard device \citep{ref34} for the particular cohort.
We first develop a deep multitask learning (MTL) regression model 
using a version of the 
same dataset produced by \citeauthor{ref31}, but with an additional user.
This model allows for more adaptable transfer learning than an AdaBoost regression model, and focuses on a beat-to-beat blood pressure estimation task as a new baseline.

One problem in training this deep neural network is that it requires a great amount of training data. Previous work on this dataset uses 80\% of all available data, over 10 minutes on average, from each subject to train the personal models \citep{ref31}. However, this calibration period is burdensome and the goal of an independent device should be to minimize the amount of calibration time required to improve utility and align with clinical need.  To that end, in this work we investigate techniques to reduce the amount of data involved in training a model. Directly training an MTL model on reduced training data fails with errors exceeding ISO standards. 
Therefore, to meet the ISO standards\footnote{ISO standards need to be met for multiple cohorts which should be representative of different populations.  In this work, only one cohort is studied.  For the sake of brevity when referring to ISO standards we refer specifically to the studied cohort and not to future cohorts.}
(in this cohort) while minimizing training data, we must utilize a technique to learn from other subjects.
A trivial way to accomplish this task would appear to be to build a generalized model from many other subjects.  However, due to inter-subject variability, generalized models we trained in leave-one-subject-out approaches produce errors greater than 10 mmHg, beyond the limits of the ISO standard. Transfer learning from a pretrained model is another solution, but the difference between subjects still impedes the learning process. 
Domain adaption \citep{ref55,ref56,ref57} is one solution to cross-domain problems, and has recently been applied with deep learning techniques \citep{ref64,ref65} to minimize the maximum mean discrepancy distance between disparate outputs.
Domain-adversarial neural networks (DANN) \citep{ref37} allows for using adversarial training to extract domain-invariant features, allowing for rapid model adaptation with minimal training data.

In this paper, we propose a DANN-based MTL model to estimate beat-to-beat blood pressure for the goal of maintaining accuracy to within ISO standards while minimizing the amount of required training data \citep{ref37}. 
To maximize clinical utility, we aim to train this model with a maximum of five minutes of training data for a new user.
Our base model, an MTL blood pressure (BP) estimation model, is composed of a long short-term memory (LSTM) coupled to a shared dense layer to extract heart beat features, and then two task-specific networks, one each for estimating diastolic and systolic blood pressure. When applying DANN, a domain (subject) classifier then attempts to classify a given beat as belonging to a particular subject. The adversarial training approach is then applied to this system with the goal of maximizing the performance of the BP estimator while minimizing the performance of the domain classifier.  Throughout this process the BP estimator is trained with reduced data from the new subject until convergence is achieved.

\subsection*{Generalizable Insights about Machine Learning in the Context of Healthcare}
\label{sec:intro:gen}

\begin{itemize}

\item Transfer learning can fail when individuals are very different.  Differences in subjects (domains) make knowledge transfer or development of a generalized model infeasible as underlying physiologies are sufficiently different that no centralized representation is easily found nor an obvious transfer mechanism is apparent.

\item DANN allows for an adversarial approach to force a model to learn subject-invariant features. This approach allows for rapid model personalization with minimal new user data. 
 \end{itemize}

\section{Related Work}
\label{sec:rel}
\subsection{Cuffless Blood Pressure}

Utilizing ECG and PPG signals to predict blood pressure is an active area of research \citep{ref40, ref32}.  Similarly, some work has been done in using dual PPG systems for this predictive task \citep{ref42, ref43}.
However, all of these methods suffer from errors introduced by variable pre-ejection periods which are highly influenced by stress, emotion, physical exertion, and age \citep{ref41}.

Another approach to cuffless blood pressure estimation, proposed by \citet{ref31}, involved the use of a wrist-worn array of bioimpedance signals consisting of 4 pairs of sensors on the ulnar and radial arteries.
Variations in bioimpedance along these arteries are correlated with PTT, the time taken for the pressure pulse to travel between two points. PTT is one of the most prominent markers used for estimation of BP. By directly evaluating measurements along the wrist, this method does not suffer from timing errors as a result of the pre-ejection period.
\citet{ref31} extracted a total of 50 features from these four bioimpedance curves using four characteristic points representing diastolic peak, maximum slope, systolic foot, and inflection point before building separate AdaBoost regression models for each subject. The calculated features and reference labels were averaged over 10-beat windows with 50\% overlap to reduce the effect of beat-to-beat variability.
These models were trained using 80\% of all available data.
Their results show an average correlation coefficient (R) and RMSE of 0.77 and 2.6 mmHg for the diastolic blood pressure and 0.86 and 3.4 mmHg for the systolic blood pressure.

\subsection{Cross-Domain Model Generalization}
When individual systems in a problem have unique variations, it can become difficult to generalize a single model to provide good results across multiple individuals. Differences between domains may result from different root causes. For instance, different data collection methods \citep{ref51}, different equipment settings \citep{ref52}, different learning tasks \citep{ref53}, or unusual frequency of observed events \citep{ref54}, can each lead to systems that superficially appear similar, and yet are unable to easily be generalized by a single model. Domain adaption techniques are one traditional solution applied for this type of cross-domain problem \citep{ref55,ref56,ref57}. Another solution involves knowledge transfer, which transfers models between domains by reweighing probability distributions among instances \citep{ref58,ref59}, clustering trained source domains \citep{ref60,ref61}, or feature transformation \citep{ref62,ref63}. Many recent deep learning-based domain adaption techniques minimize the maximum mean discrepancy distance between the outputs \citep{ref64,ref65}. However, these techniques focus on learning explicitly the difference or transformation between domains, and do not focus on learning a domain-agnostic representation. 

To address the cross-domain problem in inter-subject datasets, it is beneficial to learn useful knowledge representations while discarding subject-specific features that may hinder performance of the overall task.
Domain-adversarial neural network (DANN) \citep{ref37}, inspired by domain adaptation theory, uses an adversarial training approach to extract domain-invariant features. DANN has three parts: a feature extractor, a label predictor, and an additional domain classifier. A gradient reversal layer is applied to the feature extractor, targeting the loss from domain classification in order to adversarially train the feature extractor and domain classifier. In this way, the feature captured from source and target domains can be represented in a domain-invariant manner. DANN is widely applied in natural language processing and computer vision \citep{ref66,ref67,ref68}, solving the cross-domain transfer learning problem. \citet{ref69} proposed a pore detection method from cross-sensor fingerprint images. \citet{ref51} applied DANN in order to transfer knowledge of subject morbidity statistics from the UK Biobank into a smartphone-based HK dataset with drastically different demographic and life-style distributions, building a novel health risk model based on intraday physical activities. \citet{ref70} and \citet{ref71} applied DANN on human activity and gesture recognition in order to account for unlabeled data and \citet{ref72} extended the domain adaption algorithm to select disparate sources to use in activity recognition. 

\section{Adversarial Learning for Blood Pressure Estimation with Reduced Training Data}
\label{sec:dann}

We describe our dataset and its preprocessing in Section 3.1.  We design a baseline model for estimating blood pressure using 80\% of the data to train, detailed in Section 3.2. Reducing the size of this dataset is an important clinical challenge. We introduce DANN in Section 3.3 to accomplish this reduction. This model is adversarial \citep{ref37}, using a min-max optimization between the domain classifier and feature extractor: training the domain classifier for higher accuracy while minimizing the feature extractor to have low domain classification accuracy results in blinding the final model domains, forcing the model to rely on user-invariant features.

\subsection{Dataset and Data Preprocessing}
The BP dataset was collected using a wrist bioimpedance sensor \citep{ref31} on 11 subjects. The sensor uses four channels to measure the impedance of skin surface moments at the ulnar and radial arteries. Each subject performed a variety of physical activities in order to achieve a range of blood pressures, rising and falling with exercise and rest. A Finapres NOVA device was used at the same time to capture beat-to-beat diastolic and systolic blood pressure as ground truth reference measurements. The signals were segmented by heartbeat and samples corrupted by motion artifacts or non-physically realistic values were removed.
We downsampled the signals from the original 20 kHz to 100 Hz by equally sampling and applied zero padding to the beginning of each sequence. The first derivative of the four signal channels and the timing of each point are augmented as additional features, resulting in 9 input features in total.
During training, all training features and labels are normalized between 0 and 1.  Test features and labels were then scaled by factors learned from training normalization.

\subsection{MTL BP Estimation Model}
The base model consists of an LSTM layer, a shared dense layer, and two task-specific networks. The heartbeat data (and associated derived channels) are sent to an LSTM layer and then on to a shared dense layer. LSTM can memorize historical information, and therefore is applied to capture the patterns in the signal over time. The ability of an LSTM to retain historical information is valuable across the entirety of the heartbeat. We include a dropout layer following the LSTM. This layer allows the model to avoid overfitting and permits for some robustness to noise. Even if a part of the signal is corrupted, the model will still be able to perform with reasonable accuracy. We add a shared layer after the LSTM to further extract the relational information between channels. The extracted features are then passed on to the BP estimation network, consisting of two separate task-specific networks to estimate diastolic and systolic blood pressures.
After each layer in these two task-specific networks, a dropout layer is applied to avoid overfitting.
In order to build models for new subjects with reduced data, we further propose using DANN to transfer knowledge from other subjects and focus our attention on the beat-to-beat model as it provides higher potential clinicial utility.

\subsection{Adversarial Training with Minimal Data}

With enough data from a subject we are able to build a blood pressure regression model for that subject to within ISO standards.
However, it is desirable to improve upon this and discover the minimal amount of training data that can provide for blood pressure while remaining within the ISO standard. For a device to be implemented in clinical practice, it should be widely adaptable to a variety of patients with minimal calibration time. Therefore, our objective here is to push the limits of training data utilized while remaining precise to the necessary standards.

When simply training with less data, the model quickly produces erroneous estimates that fall out of ISO standards after a small reduction in training data. To address this issue, we investigate transfer learning solutions to more rapidly adapt our model to a previously unknown subject. However, a chief challenge of model adaptation is that the difference in wearable sensor signal data between individuals is too large, and a single generalized model fails. Therefore, we need to learn from other subjects but discard the difference between subjects.

\begin{figure}
\centering
\includegraphics[width=0.75\textwidth]{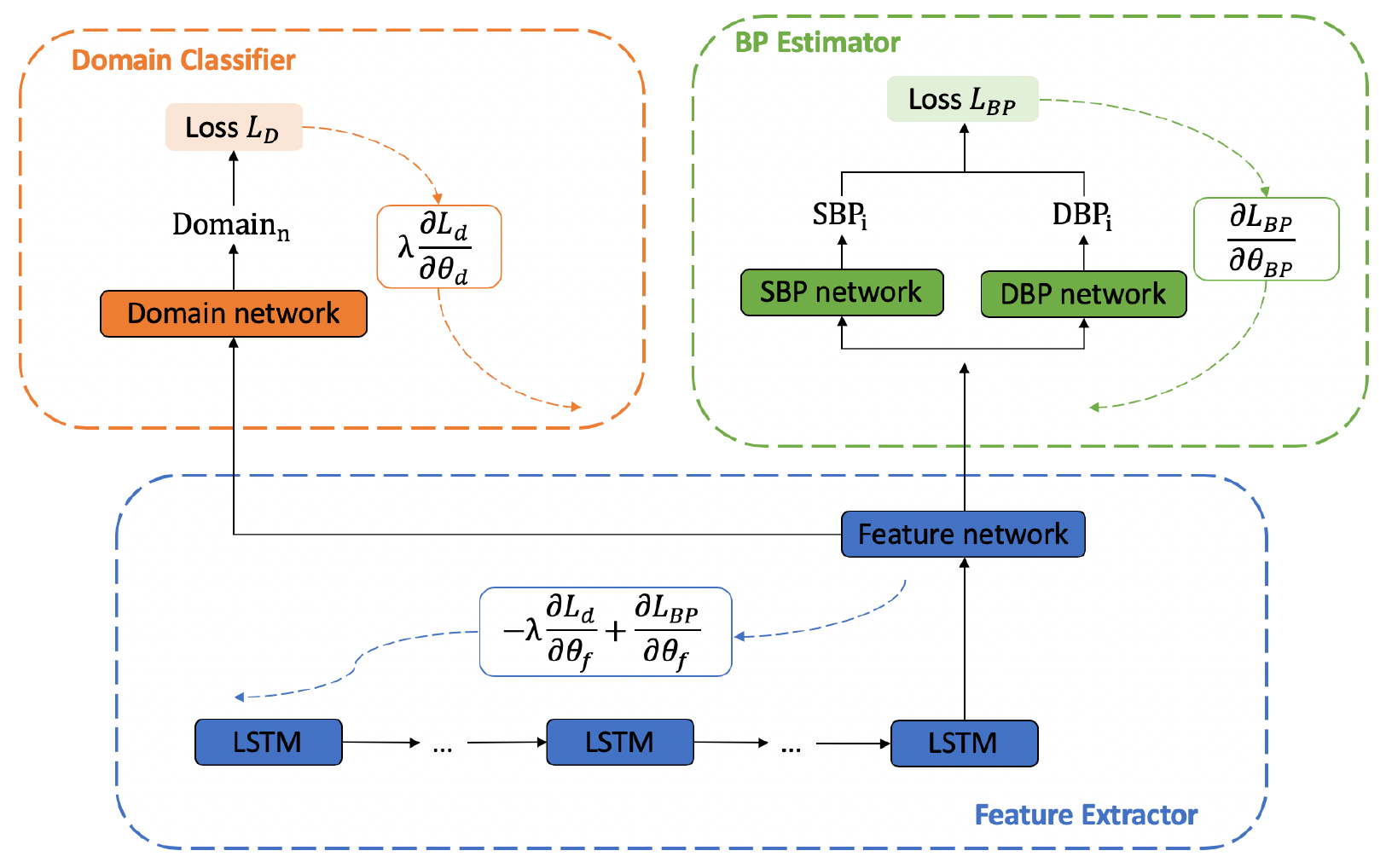}
\caption{Adversarial training structure. There are three components: Feature extractor (blue), BP estimator (green), and Domain classifier (orange). The black solid lines represent data and arrows with dashed lines represent the Systolic and Diastolic loss, respectively, for gradient descent.}\label{fig:fig7}
\end{figure} 

To build models for new subjects with reduced data, we utilize DANN to extract user-invariant features for the purpose of knowledge transfer. Figure~\ref{fig:fig7} shows the implementation of DANN within our MTL model. Our DANN model has three key parts: a feature extractor, a BP estimator, and a domain classifier. The feature extractor and BP estimator are as described above: the feature extractor is an LSTM and the BP estimator is two task-specific networks. 
The domain classifier is described in detail below and serves as a new module that pushes learning of subject-agnostic features of the data.

We treat each subject as an individual domain.
The domain classifier is trained to maximize its accuracy in recognizing to which subject a beat belongs.  The BP estimator is trained to maximize the BP regression accuracy.  The feature extractor is trained using each of these losses, but the gradient is reversed for the domain classification.  This gradient reversal pushes the feature extractor to be blind to subjects, causing the extracted features to be subject-invariant.
This coupling of the BP regression with reversed domain classification is the key adversarial component of this model.
The BP estimator, domain classifier, and feature extractor are updated using back propagation as follows:
\begin{equation}
    \theta_{BP} = \theta_{BP} + \alpha \cdot \frac{\partial L_{BP}}{\partial \theta_{BP}}
\end{equation}
\begin{equation}
    \theta_{d} = \theta_{d} + \alpha \cdot \lambda \cdot \frac{\partial L_{d}}{\partial \theta_{d}}
\end{equation}
\begin{equation}
    \theta_{f} = \theta_{f} + \alpha \cdot \left( -\lambda \cdot \frac{\partial L_{d}}{\partial \theta_{f}} + \frac{\partial L_{BP}}{\partial \theta_{f}} \right)
\end{equation}
Here $\theta$ refers the parameters in a model: $\theta_{BP}$, $\theta_d$ and $\theta_f$ indicate the parameters in the BP estimator, domain classifier, and feature extractor, respectively. $L_{BP}$ is the loss of the BP estimator and $L_d$ is the cross-entropy loss from the domain classification. $\alpha$ is the learning rate, and $\lambda$ is the loss weight, which balances the BP estimator and the domain classifier and is set to 1 in our experiments. $L_{BP}$ is given as
\begin{equation}
    L_{BP} = \sum_i \left( (E^S_i - T^S_i)^2 + (E^D_i - T^D_i)^2 \right)
\end{equation}
\noindent where $E^D$ and $E^S$ represent the estimated diastolic and systolic pressures, respectively, and $T^S$ and $T^D$ are their target values. This loss function ensures that both feature regression networks are related. Using an adversarial training approach, the feature extractor is trained to be blind to the source of the samples. Using DANN, we try to discriminate the difference between subjects, and lead the feature extractor to obtain common information that is related to blood pressure among different subjects, so that the new subject can learn from other subjects with greater training data.

Initially, we use the new subject with reduced data as our target domain, and randomly choose another subject as source domain. However, in this case the domain classifier always predicts the domain to be the source. This results from the unbalanced data between source and target domains, and any actions to the domain classifier result in a decrease of temporal accuracy. Therefore, the domain classifier stays in the local minimum and can not be updated further. To solve this problem, we introduce a second training subject as the target domain which guides the network being trained toward the new subject. We select the subjects randomly because of a lack of feasible subject similarity metric, and discuss this limitation further in Section~\ref{sec:dis:lim}. After training DANN to have stable loss, we use the reduced training data from the new subject again to retrain the model, converting the obtained knowledge from the other two subjects to align better with the new subject. Finally, we train a model under a leave-one-subject-out scheme where all other subjects are used to train the DANN. However, this approach does not converge and no usable results are produced. 

\subsection{Experiments}
To test the model performance with the reduced data, we initially limit training data to three minutes for each subject, using the remaining data as the test set. Three minutes was selected as a length of time that would be feasible for in-clinic calibration of the blood pressure system.  We first train the model directly without any pretrained model loaded or technique applied during training, so that we can understand the performance from the limited training data. Then, in order to learn from other subjects, we load the pretrained model with 80\% training data, and retrain the model with the reduced training set from the new subject. All layers of the pretrained model are retrained to adapt both the feature extraction and BP estimation functions to the new subject. For each subject, we test the pretraining approach from all the other subjects individually, and calculate the average RMSE and correlation. To evaluate the DANN model, we need two other subjects as source domain and target domain for the adversarial training approach other than the new subject. These two subjects are randomly picked from all other subjects, and we run the test 10 times for each subject as a new subject for robustness. The average RMSE and correlation are calculated as well after the 10 rounds of testing. We use the same model structure and hyperparameters in these experiments: three layers of task-specific networks with hidden size 30, learned from manually trained MTL models. This work is implemented in Python 3.6 with Tensorflow 1.15, Numpy 1.18, sklearn 0.21. The average computation time is 8.5 $\pm$ 0.5 minutes per subject without additional parallelization or fine-tuning on our server of 2 Xeon 2.2GHz CPUs, 8 GTX 1080ti GPUs, and 528 GB RAM.  
Code for this implementation can be found at \url{https://github.com/stmilab/cufflessbp_dann}.

The results of training with three minutes are not sufficient to reach ISO standards with this model. Therefore, we repeat these experiments with four and five minutes of training data. After analysis of training with four minutes, the DANN model performs within the ISO standards of 85\% of all diastolic and systolic data points having less than 10 mmHg absolute error within this cohort.

The results of utilizing only three minutes of training data are shown in table~\ref{tab:3min}, and table~\ref{tab:4min} and table~\ref{tab:5min} show results utilizing four and five minutes of training data respectively. From the results, using three minutes of training data obtains an RMSE of $4.80 \pm 0.74$ mmHg for diastolic blood pressure and $7.34 \pm 1.88$ mmHg for systolic blood pressure. DANN improves RMSE over the pretrained model by 0.20 mmHg for diastolic blood pressure and 0.60 mmHg for systolic blood pressure. When utilizing four minutes of training data, the model obtains an RMSE of $4.64 \pm 0.60$ mmHg for diastolic blood pressure and $7.10 \pm 1.79$ mmHg for systolic blood pressure. DANN improves RMSE over the pretrained model by 0.19 mmHg for diastolic blood pressure and 0.46 mmHg for systolic blood pressure. For five minutes training data, DANN improves RMSE over the pretrained model by 0.26 mmHg to 4.48 mmHg for diastolic blood pressure, and improves by 0.67 mmHg to  6.79 mmHg for systolic blood pressure\footnote{Subject 6 has only five minutes of data in total, and so is excluded from analyses with five minutes training data.}. We also test the pretrained models on new users without any retraining process. The average RMSE for DBP is 6.94 mmHg and for SBP is 11.51 mmHg, and the average correlation for DBP is 0.07 and for SBP is -0.01.

\begin{table}[t]
  \footnotesize
  \centering 
  \caption{Results using three minutes of subject-specific training data for diastolic and systolic blood pressure (DBP \& SBP)}
  \begin{tabular}{cccccccc}
    \multicolumn{2}{c}{Subject} & \multicolumn{2}{c}{DANN} & \multicolumn{2}{c}{Pretrained} & \multicolumn{2}{c}{Directly Trained} \\ \hline
 & & RMSE & R & RMSE & R & RMSE & R    \\
    \hline
1&	DBP:&	$\mathbf{4.56 \pm 0.07}$& $\mathbf{0.43 \pm 0.05}$&	$4.93 \pm 0.14$& $0.33 \pm 0.07$& $4.93$ & $0.16$\\ 
&	SBP:& 	$\mathbf{5.98 \pm 0.06}$& $\mathbf{0.25 \pm 0.03}$&	$6.19 \pm 0.11$& $0.11 \pm 0.05$& $12.88$ &$0.00$\\

2&	DBP:&	$\mathbf{5.39 \pm 0.12}$& $\mathbf{0.57 \pm 0.03}$&	$5.72 \pm 0.14$& $0.47 \pm 0.04$& $6.44$ &$0.00$\\
&	SBP:&	$\mathbf{8.45 \pm 0.20}$& $\mathbf{0.65 \pm 0.02}$&	$9.24 \pm 0.30$& $0.55 \pm 0.04$& $12.91$ &$0.02$\\

3&	DBP:&	$\mathbf{4.08 \pm 0.11}$& $\mathbf{0.40 \pm 0.02}$&	$4.22 \pm 0.11$& $0.23 \pm 0.11$& $13.65$ &$0.00$\\
&	SBP:& 	$\mathbf{6.06 \pm 0.14}$& $\mathbf{0.50 \pm 0.03}$&	$6.81 \pm 0.19$& $0.36 \pm 0.10$& $7.41$ &$0.00$\\

4&	DBP:&	$\mathbf{4.21 \pm 0.05}$& $\mathbf{0.07 \pm 0.05}$&	$4.29 \pm 0.16$& $0.02 \pm 0.04$& $4.12$ &$0.05$\\
&	SBP:& 	$\mathbf{7.63 \pm 0.03}$& $\mathbf{0.18 \pm 0.03}$&	$8.11 \pm 0.21$& $0.16 \pm 0.07$& $17.26$ &$0.00$\\

5&	DBP:&	$\mathbf{5.15 \pm 0.07}$& $0.22 \pm 0.06$&	$5.52 \pm 0.23$& $\mathbf{0.23 \pm 0.04}$& $5.61$ &$0.20$\\
&	SBP:&	$\mathbf{5.95 \pm 0.12}$& $0.26 \pm 0.09$&	$6.22 \pm 0.24$& $\mathbf{0.28 \pm 0.02}$& $6.02$ &$0.30$\\

6&	DBP:&	$\mathbf{6.25 \pm 0.09}$& $\mathbf{0.29 \pm 0.04}$&	$6.41 \pm 0.18$& $0.23 \pm 0.04$& $7.26$ &$0.19$\\
&	SBP:&	$\mathbf{7.59 \pm 0.13}$& $\mathbf{0.55 \pm 0.02}$&	$8.16 \pm 0.24$& $0.46 \pm 0.04$& $9.16$ &$0.00$\\

7&	DBP:&	$\mathbf{5.20 \pm 0.07}$& $\mathbf{0.29 \pm 0.05}$&	$5.60 \pm 0.14$& $0.22 \pm 0.05$& $6.09$ &$0.25$\\
&	SBP:&	$\mathbf{8.21 \pm 0.06}$& $\mathbf{0.37 \pm 0.07}$&	$8.76 \pm 0.15$& $0.33 \pm 0.05$& $8.89$ &$0.00$\\

8&	DBP:&	$\mathbf{5.50 \pm 0.11}$& $\mathbf{0.27 \pm 0.10}$&	$5.77 \pm 0.13$& $0.24 \pm 0.11$& $5.74$ &$0.20$\\
&	SBP:&	$\mathbf{12.06 \pm 0.24}$& $\mathbf{0.30 \pm 0.03}$&	$12.88 \pm 0.54$& $0.30 \pm 0.09$& $12.82$ &$0.30$\\

9&	DBP:&	$\mathbf{4.02 \pm 0.06}$& $\mathbf{0.34 \pm 0.02}$&	$4.22 \pm 0.10$& $0.21 \pm 0.05$& $4.72$ &$0.21$\\
&	SBP:&	$\mathbf{5.47 \pm 0.06}$& $\mathbf{0.18 \pm 0.08}$&	$5.81 \pm 0.20$& $0.07 \pm 0.04$& $5.56$ &$0.00$\\

10&	DBP:&	$\mathbf{4.23 \pm 0.01}$& $\mathbf{0.12 \pm 0.02}$&	$4.34 \pm 0.08$& $0.12 \pm 0.05$& $4.24$ &$0.07$\\
&	SBP:&	$\mathbf{5.86 \pm 0.02}$& $\mathbf{0.17 \pm 0.02}$&	$6.00 \pm 0.10$& $0.12 \pm 0.04$& $5.93$ &$0.06$\\

11&	DBP:&	$\mathbf{4.24 \pm 0.08}$& $\mathbf{0.51 \pm 0.02}$&	$4.61 \pm 0.09$& $0.36 \pm 0.06$& $4.44$ &$0.49$\\
&	SBP:&	$\mathbf{7.42 \pm 0.18}$& $\mathbf{0.51 \pm 0.03}$&	$8.14 \pm 0.17$& $0.33 \pm 0.07$& $8.47$ &$0.00$\\
\hline

Mean&	DBP:&$\mathbf{4.80 \pm 0.74}$& $\mathbf{0.32 \pm 0.15}$& $5.06 \pm 0.78$& $0.24 \pm 0.12$& $6.11 \pm 2.56$ &$0.16 \pm 0.14$\\
&	SBP:&$\mathbf{7.34 \pm 1.88}$  &$\mathbf{0.36 \pm 0.17}$& $7.84 \pm 2.06$& $0.28 \pm 0.15$& $9.75 \pm 3.57$ & $0.06 \pm 0.12$\\

    \hline 
  \end{tabular}
  \label{tab:3min} 
\vspace*{-0.1cm}
\end{table}

\begin{table}[h]
  \footnotesize
  \centering 
  \caption{Results using four minutes of subject-specific training data for diastolic and systolic blood pressure (DBP \& SBP)}
  \begin{tabular}{cccccccc}
    \multicolumn{2}{c}{Subject} & \multicolumn{2}{c}{DANN} & \multicolumn{2}{c}{Pretrained} & \multicolumn{2}{c}{Directly Trained} \\ \hline
 & & RMSE & R & RMSE & R & RMSE & R    \\
    \hline
1&	DBP:&	$\mathbf{4.49 \pm 0.08}$&  $\mathbf{0.45 \pm 0.03}$&	$4.81 \pm 0.11$& $0.32 \pm 0.08$ & $5.10$ & $0.37$\\ 
&	SBP:& 	$\mathbf{5.92 \pm 0.08}$&  $\mathbf{0.26 \pm 0.05}$&	$6.12 \pm 0.10$& $0.12 \pm 0.05$ & $6.20$ & $0.18$\\

2&	DBP:& 	$\mathbf{5.32 \pm 0.11}$&  $\mathbf{0.58 \pm 0.03}$&	$5.60 \pm 0.15$& $0.51 \pm 0.04$ & $5.36$ & $0.57$\\
&	SBP:& 	$\mathbf{8.19 \pm 0.29}$&  $\mathbf{0.68 \pm 0.03}$&	$9.12 \pm 0.30$& $0.58 \pm 0.04$ & $8.90$ & $0.62$\\

3&	DBP:& 	$\mathbf{3.96 \pm 0.06}$&  $\mathbf{0.42 \pm 0.03}$&	$4.16 \pm 0.10$& $0.23 \pm 0.12$ & $4.18$ & $0.36$\\
&	SBP:& 	$\mathbf{6.03 \pm 0.28}$&  $\mathbf{0.57 \pm 0.05}$&	$6.60 \pm 0.29$& $0.43 \pm 0.09$& $7.30$ & $0.00$\\

4&	DBP:& 	$\mathbf{4.06 \pm 0.06}$&  $\mathbf{0.09 \pm 0.02}$&	$4.07 \pm 0.05$& $0.06 \pm 0.06$& $4.44$ & $0.05$\\
&	SBP:& 	$\mathbf{7.68 \pm 0.14}$&  $\mathbf{0.25 \pm 0.05}$&	$7.96 \pm 0.21$& $0.17 \pm 0.10$& $8.26$ & $0.21$\\

5&	DBP:& 	$\mathbf{5.03 \pm 0.18}$&  $0.23 \pm 0.25$&	$5.01 \pm 0.12$& $0.21 \pm 0.04$& $5.08$ & $\mathbf{0.28}$\\
&	SBP:& 	$\mathbf{5.77 \pm 0.09}$&  $\mathbf{0.28 \pm 0.03}$&	$5.82 \pm 0.14$& $0.26 \pm 0.06$& $6.20$ & $0.00$\\

6&	DBP:& 	$\mathbf{5.34 \pm 0.23}$&  $\mathbf{0.33 \pm 0.09}$&	$5.74 \pm 0.06$& $0.20 \pm 0.04$& $5.32$ & $0.30$\\
&	SBP:& 	$\mathbf{6.30 \pm 0.19}$&  $\mathbf{0.63 \pm 0.03}$&	$7.46 \pm 0.39$& $0.53 \pm 0.07$& $8.47$ & $0.39$\\

7&	DBP:& 	$\mathbf{5.17 \pm 0.10}$&  $0.33 \pm 0.27$&	$5.24 \pm 0.11$& $0.26 \pm 0.08$& $5.77$ & $\mathbf{0.29}$\\
&	SBP:& 	$\mathbf{8.12 \pm 0.16}$&  $\mathbf{0.43 \pm 0.04}$&	$8.41 \pm 0.18$& $0.34 \pm 0.06$& $9.01$ & $0.40$\\

8&	DBP:& 	$\mathbf{5.34 \pm 0.12}$&  $\mathbf{0.39 \pm 0.05}$&	$5.50 \pm 0.15$& $0.34 \pm 0.08$& $5.35$ & $0.38$\\
&	SBP:& 	$\mathbf{11.62 \pm 0.34}$&  $\mathbf{0.44 \pm 0.06}$&	$12.07 \pm 0.25$& $0.38 \pm 0.07$& $12.14$ & $0.34$\\

9&	DBP:& 	$\mathbf{3.98 \pm 0.07}$&  $\mathbf{0.33 \pm 0.07}$&	$4.18 \pm 0.06$& $0.21 \pm 0.06$& $4.15$ & $0.28$\\
&	SBP:& 	$\mathbf{5.47 \pm 0.08}$&  $\mathbf{0.24 \pm 0.04}$&	$5.68 \pm 0.08$& $0.06 \pm 0.04$& $5.68$ & $0.13$\\

10&	DBP:& 	$\mathbf{4.19 \pm 0.03}$&  $0.12 \pm 0.04$&	$4.25 \pm 0.07$& $\mathbf{0.13 \pm 0.03}$& $4.68$ & $0.07$\\
&	SBP:& 	$\mathbf{5.83 \pm 0.02}$&  $\mathbf{0.13 \pm 0.02}$&	$5.90 \pm 0.09$& $0.13 \pm 0.04$& $6.43$ & $0.04$\\

11&	DBP:& 	$\mathbf{4.15 \pm 0.06}$&  $\mathbf{0.52 \pm 0.06}$&	$4.54 \pm 0.12$& $0.38 \pm 0.08$& $4.56$ & $0.38$\\
&	SBP:& 	$\mathbf{7.25 \pm 0.13}$&  $\mathbf{0.52 \pm 0.02}$&	$7.99 \pm 0.16$& $0.37 \pm 0.08$& $8.07$ & $0.42$\\

\hline

Mean&	DBP:& 	$\mathbf{4.64 \pm 0.60}$&  $\mathbf{0.34 \pm 0.15}$&	$4.83 \pm 0.62$& $0.26 \pm 0.12$ & $4.90 \pm 0.53$ & $0.31 \pm 0.15$\\
&	SBP:& 	$\mathbf{7.10 \pm 1.79}$& $\mathbf{0.40 \pm 0.18}$&   $7.56 \pm 1.90$&	$0.31 \pm 0.17$& $7.88 \pm1.84$ & $0.25 \pm 0.20$\\

    \hline 
  \end{tabular}
  \label{tab:4min} 
\vspace*{-0.1cm}
\end{table}


\begin{table}[h]
  \footnotesize
  \centering 
  \caption{Results using five minutes of subject-specific training data for diastolic and systolic blood pressure (DBP \& SBP)}
  \begin{tabular}{cccccccc}
    \multicolumn{2}{c}{Subject} & \multicolumn{2}{c}{DANN} & \multicolumn{2}{c}{Pretrained} & \multicolumn{2}{c}{Directly Trained} \\ \hline
 & & RMSE & R & RMSE & R & RMSE & R    \\
    \hline
1&	DBP:&	$\mathbf{4.39 \pm 0.07}$& $\mathbf{0.45 \pm 0.04}$& $4.79 \pm 0.16$& $0.34 \pm 0.09$& $4.87$ &$0.37$\\
&	SBP:&	$\mathbf{5.85 \pm 0.08}$& $\mathbf{0.38 \pm 0.03}$& $6.04 \pm 0.07$& $0.15 \pm 0.06$& $6.03$ &$0.00$\\

2&	DBP:&	$\mathbf{5.26 \pm 0.07}$& $\mathbf{0.59 \pm 0.03}$& $5.63 \pm 0.10$& $0.51 \pm 0.02$& $5.34$ &$0.56$\\
&	SBP:& 	$\mathbf{7.98 \pm 0.17}$& $\mathbf{0.68 \pm 0.02}$& $9.00 \pm 0.15$& $0.60 \pm 0.02$& $8.58$ &$0.66$\\

3&	DBP:&	$\mathbf{3.89 \pm 0.06}$& $\mathbf{0.44 \pm 0.03}$& $4.13 \pm 0.13$& $0.23 \pm 0.15$& $3.90$ &$0.44$\\
&	SBP:& 	$\mathbf{5.77 \pm 0.15}$& $\mathbf{0.58 \pm 0.02}$& $6.26 \pm 0.37$& $0.51 \pm 0.08$& $7.19$ &$0.00$\\

4&	DBP:&	$4.06 \pm 0.04$& $\mathbf{0.10 \pm 0.02}$& $\mathbf{4.04 \pm 0.06}$& $0.03 \pm 0.07$& $4.37$ &$0.11$\\
&	SBP:& 	$\mathbf{7.69 \pm 0.06}$& $\mathbf{0.29 \pm 0.04}$& $8.02 \pm 0.18$& $0.20 \pm 0.10$& $7.87$ &$0.23$\\

5&	DBP:&	$\mathbf{4.83 \pm 0.29}$& $\mathbf{0.26 \pm 0.07}$& $4.87 \pm 0.14$& $0.18 \pm 0.10$& $5.11$ &$0.25$\\
&	SBP:& 	$\mathbf{5.61 \pm 0.06}$& $\mathbf{0.27 \pm 0.04}$& $5.85 \pm 0.18$& $0.21 \pm 0.08$& $6.05$ &$0.19$\\
6&-&-&-&-&-&-&-\\
&-&-&-&-&-&-&-\\
7&	DBP:&	$\mathbf{5.04 \pm 0.05}$& $\mathbf{0.36 \pm 0.02}$& $5.21 \pm 0.08$& $0.32 \pm 0.03$& $5.61$ &$0.30$\\
&	SBP:&	$\mathbf{7.94 \pm 0.06}$& $\mathbf{0.47 \pm 0.02}$& $8.28 \pm 0.16$& $0.41 \pm 0.04$& $8.69$ &$0.43$\\

8&	DBP:&	$\mathbf{5.27 \pm 0.16}$& $\mathbf{0.40 \pm 0.03}$& $5.50 \pm 0.21$& $0.34 \pm 0.11$& $5.49$ &$0.37$\\
&	SBP:& 	$\mathbf{10.83 \pm 0.39}$& $\mathbf{0.48 \pm 0.05}$& $12.04 \pm 0.57$& $0.39 \pm 0.12$& $12.98$ &$0.35$\\

9&	DBP:&	$\mathbf{3.84 \pm 0.06}$& $\mathbf{0.34 \pm 0.03}$& $4.04 \pm 0.10$& $0.23 \pm 0.07$& $4.29$ &$0.24$\\
&	SBP:&	$\mathbf{5.29 \pm 0.04}$& $\mathbf{0.29 \pm 0.03}$& $5.53 \pm 0.11$& $0.09 \pm 0.03$& $5.63$ &$0.03$\\

10&	DBP:&	$\mathbf{4.20 \pm 0.02}$& $0.13 \pm 0.03$& $4.31 \pm 0.09$& $0.11 \pm 0.05$& $4.44$ &$\mathbf{0.18}$\\
&	SBP:& 	$\mathbf{5.87 \pm 0.02}$& $\mathbf{0.18 \pm 0.04}$& $5.98 \pm 0.06$& $0.13 \pm 0.05$& $6.32$ &$0.16$\\

11&	DBP:&	$\mathbf{4.02 \pm 0.05}$& $0.53 \pm 0.02$& $4.30 \pm 0.16$& $0.48 \pm 0.05$& $4.13$ &$\mathbf{0.54}$\\
&	SBP:& 	$\mathbf{6.91 \pm 0.12}$& $\mathbf{0.54 \pm 0.02}$& $7.62 \pm 0.19$& $0.45 \pm 0.05$& $8.48$ &$0.48$\\

\hline
Mean&	DBP:&$\mathbf{4.48 \pm 0.57}$& $\mathbf{0.36 \pm 0.16}$& $4.68 \pm 0.60$& $0.28 \pm 0.15$& $4.76 \pm 0.58$ &$0.33 \pm 0.14$\\
&SBP:& $\mathbf{6.79 \pm 1.70}$& $\mathbf{0.41 \pm 0.17}$& $7.46 \pm 2.00$& $0.31 \pm 0.18$& $7.78 \pm 2.05$ & $0.25 \pm 0.22$\\

    \hline 
  \end{tabular}
  \label{tab:5min} 
\vspace*{-0.1cm}
\end{table}

\begin{figure}[h]
\centering
\includegraphics[width=0.7\textwidth]{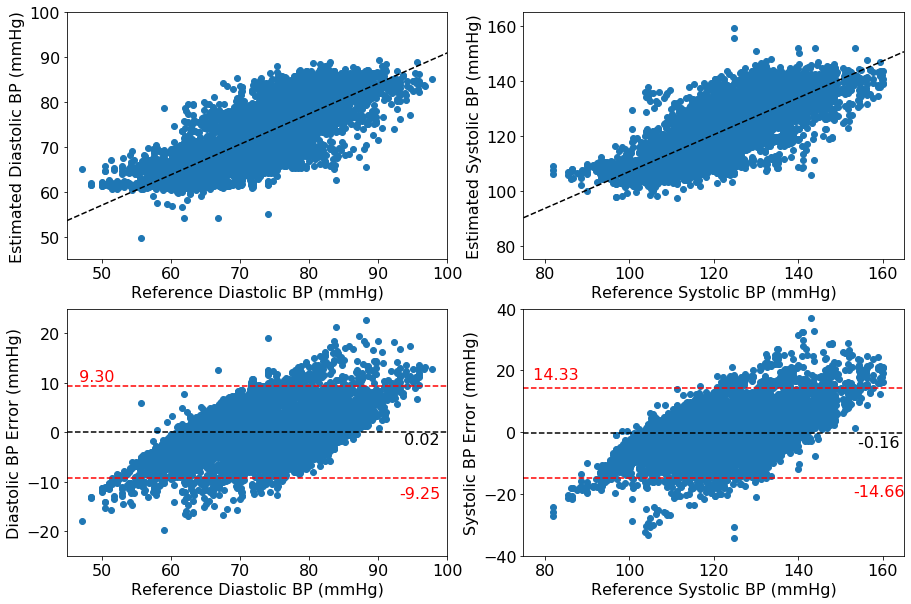}
\caption{Bland-Altman plot for DANN model using three minutes of subject-specific training data}\label{fig:fig8}
\vspace*{-0.2cm}
\end{figure}

\begin{figure}[h]
\centering
\includegraphics[width=0.7\textwidth]{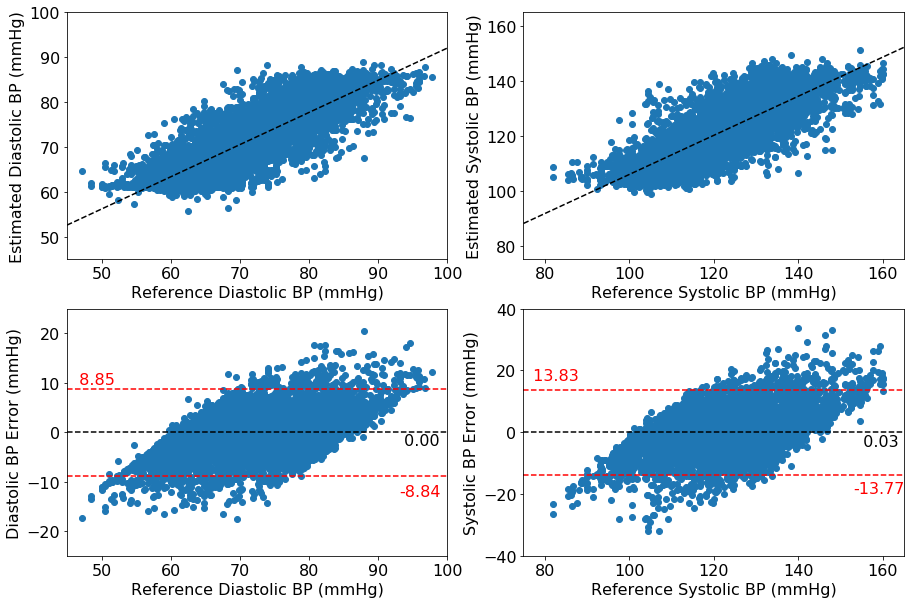}
\caption{Bland-Altman plot for DANN model using four minutes of subject-specific training data}\label{fig:ba_4min}
\vspace*{-0.2cm}
\end{figure}

\begin{figure}[h]
\centering
\includegraphics[width=0.7\textwidth]{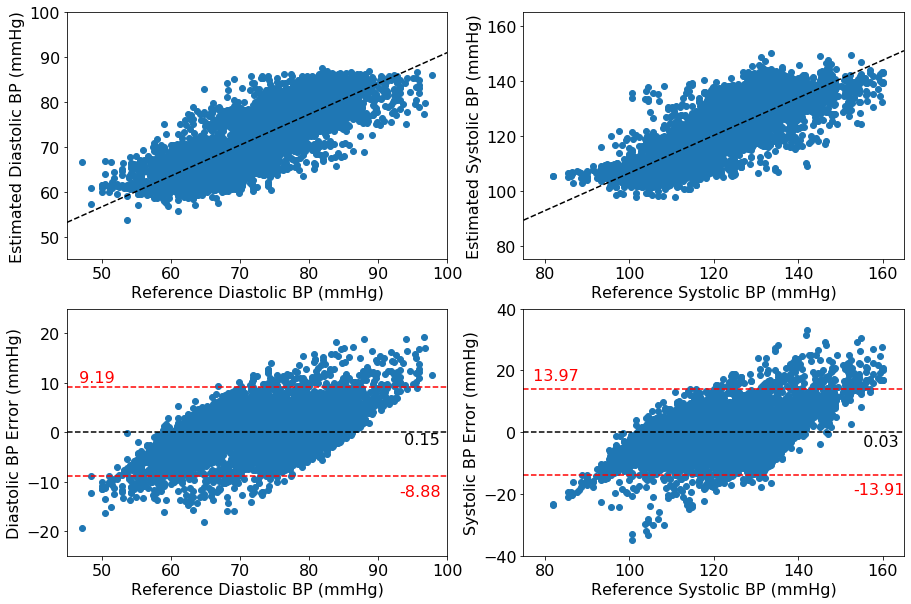}
\caption{Bland-Altman plot for DANN model using five minutes of subject-specific training data}\label{fig:fig9}
\vspace*{-0.2cm}
\end{figure}

Figures~\ref{fig:fig8}, \ref{fig:ba_4min}, and \ref{fig:fig9} show the Bland-Altman plots for the DANN model trained with three, four, or five minutes of training data, respectively. In the three-minute model, 96.0\% of predictions have diastolic error less than 10 mmHg, however only 84.5\% of predictions have systolic error less than 10 mmHg. This result is below the ISO standard, and prompt repeating the experiment with five minutes of training data. In the four-minute and five-minute model, this error improves to be within the ISO standard: 96.2\% diastolic error and 85.9\% systolic error are less than 10 mmHg in four-minute model, and 96.2\% diastolic error and 85.5\% systolic error are less than 10 mmHg in five-minute model. The decrease from the four-minute model to the five-minute model might be the missing subject in the five-minute model. 

\begin{figure}[h]
\centering
\includegraphics[width=0.4\textwidth]{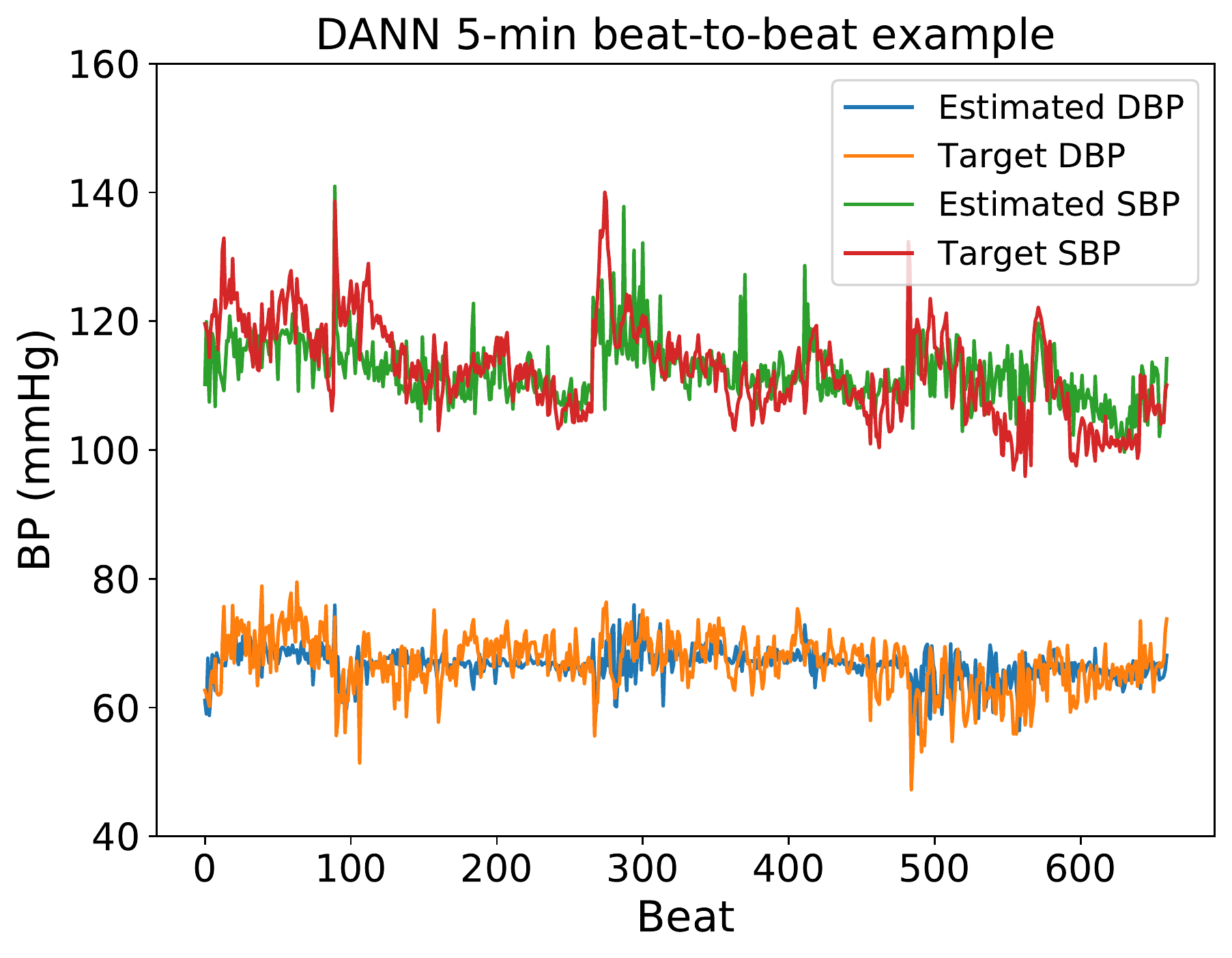}
\caption{Estimated and target blood pressure plots from a subject. The estimation here is provided by the DANN model and trained with five minutes of training data. This plot is not completely representative: for some subjects with lower variability, the model does not respond to changes in blood pressure and instead predicts a near constant blood pressure.}\label{fig:fig10}
\vspace*{-0.2cm}
\end{figure}



Figure~\ref{fig:fig10} is an example of estimated and target blood pressure with five minutes of training data applying DANN. Five minutes of training data is able to track the change of blood pressure, e.g. the systolic blood pressure in the figure. However, the reduced training data cannot always respond to changes in blood pressure, especially for cases with lower variability such as the estimation of diastolic blood pressure in the figure.

\subsection{Analysis}
From these results, we observe that model performance decreases significantly when reducing the training data, and less training data results in much lower accuracy (higher RMSE). With three minutes of training data the original MTL model without a pretrained model or the adversarial training process fails for many subjects. There are five subjects with RMSE over 10 mmHg, which is outside of the acceptable range for ISO standards in blood pressure. However, when training with a pretrained model from another subject, the model performance improves for both estimations. When applying the DANN-based training method, the RMSE further decreases, particularly for systolic blood pressure more so than for diastolic.

When training with four or five minutes of data, all three training approaches show an increase in performance. It is interesting to note that 10 out of 11 subjects obtain RMSE below 10 mmHg when directly training the MTL model without DANN. Comparing to training with the pretrained model and direct training approaches, our DANN-based model still has significant benefits. The DANN-based model has lower RMSE, lower standard deviation, and higher correlation, meaning that it performs better and more robustly for additional subjects. In comparison to direct training of the MTL model, both the DANN-based model and the pretrained model help improve the model performance, meaning that it is important to learn from other subjects when a new subject does not have enough training data. The advantage of the DANN-based model indicates that it is more useful to learn from other subjects and to discard the difference between subjects. 

With less training data, the model tends
to estimate blood pressure as closer to mean values, causing significant errors for extreme high and low blood pressure. When training the model with three minutes of data, the 85\% absolute error for systolic blood pressure is 10.11 mmHg. The 85\% error is greater than 10 mmHg of the ISO standard, even though it is already improved by applying DANN. Then, we extend the training data to meet the ISO standard requirement. In this model, four minutes of training data is the minimum required amount of training data to obtain confident blood pressure estimations within ISO standards, and while maximizing clinical convenience for future use.

\section{Discussion} 
\label{sec:dis}

In the problem of minimizing patient training data for rapid adaptation of a blood pressure prediction model, this work shows the benefits of using DANN to transfer knowledge from other subjects. Considering the difference of physical signals between subjects, rather than building a generalized model or finding a mapping from one subject to another for transformation, DANN provides an approaching of using adversarial training to extract subject-invariant features for the purpose of transfer learning. When applying DANN to this MTL model, the shared layers can be treated as the feature extractor, and the task-specific networks are the label predictor.
We have shown that this approach obtains improvements of RMSE 0.19 to 0.26 mmHg for diastolic blood pressure and 0.46 to 0.67 mmHg for systolic blood pressure in the comparison to our best baseline model. 
The adversarial training mechanism between feature extractor and domain predictor causes the shared layers in the MTL model to extract information which focus on blood pressure estimation while being trained to be insensitive to differences between subjects.  In this manner, applying the model to a previously unknown subject allows for improved training while minimizing training data. We see here that the domain predictor is an important factor in the training process. The unbalanced dataset can cause failure of domain prediction by predicting every sample to the same class, collapsing the model to lack subject-specificity. Introducing the third subject solves this problem by balancing between source and target domains.
DANN demonstrates strong performance in adaptation of the model to new individuals with minimal data. Notably, the model trained with DANN is less prone to memorizing individual subject features as compared to the isolated MTL model. We conduct additional experiments, evaluating the model's ability to interpolate missing blood pressure values, demonstrating the more general regression DANN finds, by intentionally withholding certain blood pressure ranges from the training phase and find that the RMSE of the test set for values within the training set versus those within the generated gap are similar. This additional analysis is discussed in the Supplementary Appendix.
The minimal training data for blood pressure with bioimpedance signals is four minutes in this system with the given dataset.
This provides a guidance for future data collection to validate the device on more people and with a variety of clinical conditions. Our proposed model also provides a solution to the problem of modeling physical measurement tasks with reduced training data as well as a method of learning the minimum required amount of training data.

\subsection{Limitations \& Future Directions}
\label{sec:dis:lim}
%
%

One chief limitation is the lack of understanding here of the feature space in which subject variability exists.  Given an arbitrarily large number of subjects, it is likely that similarities can be found to allow for more intelligent selection of subjects within the model.  Therefore, a chief future direction will be to develop a more formalized understanding of this space to allow for a greater ability to choose which existing training sets should be utilized in adapting the model to a new subject. While DANN as presented in this paper is utilized with two random training subjects, this technique can be adapted to incorporate data from additional subjects. One approach will be to utilize a distance metric on demographics or physiological data to base the training on the most similar subjects. Although DANN improves the model performance by transfer learning from other subjects, the correlation for some subjects is very low, and the model cannot respond well to the data with less variability. In all cases, an exploration of the types of data (e.g. low and high blood pressure values) and their relative impact in training should be further explored.

\subsection{Conclusion}
\label{sec:dis:conc}

In this paper, we propose a DANN-based MTL model to estimate beat-to-beat blood pressure from cuffless bioimpedance signals for new subjects with reduced training data. When reducing the training data to three, four, and five minutes, the base MTL model cannot directly be trained successfully to be within ISO standards. Therefore, in order to transfer knowledge from other subjects efficiently, we modify the DANN training approach to train the feature extractor for subject-invariant features. With DANN, the model obtains average RMSE $4.80 \pm 0.74$ mmHg for diastolic blood pressure and $7.34 \pm 1.88$ mmHg for systolic blood pressure when using three minutes training data, $4.64 \pm 0.60$ mmHg and $7.19 \pm 1.79$ for diastolic and systolic blood pressure from four minutes training data, and $4.48 \pm 0.57$ mmHg and $6.79 \pm 1.70$ for diastolic and systolic blood pressure, respectively, when applying five minutes training data. DANN improves the knowledge transfer ability for three, four, and five minutes of training data in comparison to directly training or training with a pretrained model from another subject, decreasing RMSE by 0.19 to 0.26 mmHg for diastolic blood pressure and by 0.46 to 0.67 mmHg for systolic blood pressure in comparison to the best baseline model of utilizing a pretrained model from another subject.
The model performance increases with additional data, and we conclude that four minutes is the minimum requirement to achieve the ISO standard with our proposed model and participant cohort. In the future, we consider expanding the DANN approach to include more subject features and to find a metric for the subject-domain space in order to choose similar subjects prior to adapting a model given new subject with targeted minimal training data.


\section*{Acknowledgements}
This work was supported, in part, by the National Institutes
of Health, under grant 1 R01 EB028106-01.

\bibliography{mlhc}

\newpage
\section*{Appendix}
\label{sec:app}

\setcounter{table}{0}
\renewcommand{\thetable}{A\arabic{table}}
\renewcommand\thefigure{A\arabic{figure}}
\setcounter{figure}{0} 

\section*{A.1 Model Performance Relative to ISO Standard}

For the development of a blood pressure device, ISO standards require that 85\% of measurements be within 10 mmHg of a standardized reference value for a given cohort.  For each subject in our dataset, Table \ref{tab:iso} reports the percentage of measurements that fall within this range for varying lengths (3 minutes, 4 minutes, or 5 minutes) of training data.  The mean values are reported as well, showing that with 4 minutes of training data, 96.1\% of DBP and 85.2\% of SBP measurements fall within this range for the cohort studied.  This framework presented allows for further adaptation of DANN training times as data collection from future cohorts progresses.

\begin{table}[H]
  \footnotesize
  \centering 
  \caption{The percentage (\%) of results from the DANN model that fall within 10 mmHg of the reference value.  To meet ISO standards in a given cohort, at least 85\% of measurements must fall within that range.}
  \begin{tabular}{ccccccc}
  \multicolumn{1}{c}{} & \multicolumn{2}{c}{3 mins} & \multicolumn{2}{c}{4 mins} & \multicolumn{2}{c}{5 mins} \\ \hline
Subject    & DBP & SBP & DBP & SBP & DBP & SBP \\ \hline
1& 95.7\%  & 91.3\% & 95.8\% & 91.7\% & 95.9\% & 90.7\% \\
2& 94.4\%  & 77.1\% & 93.8\% & 81.4\% & 91.9\% & 74.2\% \\
3& 98.4\%  & 91.7\% & 97.4\% & 91.4\% & 98.3\% & 93.7\% \\
4& 98.2\%  & 80.4\% & 98.7\% & 82.3\% & 98.8\% & 82.5\% \\
5& 90.0\%  & 85.3\% & 94.0\% & 91.0\% & 92.3\% & 91.3\% \\
6& 92.7\%  & 78.0\% & 96.4\% & 87.3\% & - & - \\
7& 95.2\%  & 79.5\% & 94.7\% & 81.7\% & 95.4\% & 83.2\% \\
8& 92.7\%  & 65.6\% & 92.5\% & 63.1\% & 94.2\% & 65.8\% \\
9& 99.7\%  & 89.4\% & 99.6\% & 93.5\% & 99.8\% & 95.2\% \\
10& 96.6\% & 89.4\% & 97.1\% & 89.6\% & 97.2\% & 89.4\% \\
11& 96.9\% & 82.4\% & 97.5\% & 83.9\% & 97.2\% & 85.6\% \\
\hline
Mean& 95.5\% & 83.1\% & 96.1\% & 85.2\% & 96.1\% & 85.2\% \\	

    \hline 
  \end{tabular}
  \label{tab:iso} 
\end{table}

\section*{A.2 Model Interpolation}

To further evaluate DANN's ability to generate a general regression model, which may aid in future reduction of needed training data, we test the ability of the model to interpolate blood pressures in specific ranges that are intentionally withheld from training. For each individual we adapt the model to using DANN, we first remove from all samples with either diastolic or systolic blood pressure within a specific range (for example, systolic blood pressure from 120-125 mmHg) from the training set. We then repeat model training (using 4-minutes of training data) and test on the full, held out test set.  This is analogous to the experiment with results recorded in Table \ref{tab:4min} but with a different distribution of training data, reducing the ranges of blood pressures seen from the new individual.

After training, we test the model with the full test set, which includes blood pressure values from the test individual withheld from training. We report both overall RMSE for all test data and in-gap RMSE where the test data exclusive comes from the omitted blood pressure range. Will illustrate model performance with diastolic gaps of 5 mmHg and systolic gaps of 6 mmHg. Specifically, we tested diastolic gaps of 55-60, 65-70, 70-75, 75-80, 80-85, and 85-90 mmHg, and systolic gaps of 90-96, 95-101, 100-106, 105-111, 110-116, 115-121, 120-126, 125-131, 130-136, 135-141, 140-146, and 145-151 mmHg. Due to variations between subjects, not all gaps were tested on all subjects. For instance, a subject whose systolic blood pressure never fell below 106 mmHg would not be included in a gap test for the systolic range of 100-106 mmHg. The overall RMSE and in-gap RMSEs were averaged over each subject, and those values are reported in Table \ref{tab:gap}.   We test these gaps at intervals throughout the distribution of blood pressures present. We note that, even with the errors introduced from the missing values, DANN still outperforms the other models.


As seen in Table \ref{tab:gap}, the model error tends to increase slightly in the gaps of training data. This is expected given that this model is never trained on values from within those gaps. However, the mean of the error within the gap and overall is still small, showing that the model is able to successfully interpolate to unseen values.

\begin{table}[H]
  \footnotesize
  \centering 
  \caption{Model results when trained using gaps in training data. DBP gap size is 5 mmHg and SBP gap size is 6 mmHg.  Results shown are averaged across varying gap locations as described in the text.}
  \begin{tabular}{ccccc}
     & \multicolumn{2}{c}{DBP} & \multicolumn{2}{c}{SBP}\\ \hline
    Subject & Overall RMSE & In-Gap RMSE & Overall RMSE & In-Gap RMSE \\ \hline
1&	4.64&	5.19&	5.99&	7.38\\
2&	5.61&	5.56&	8.23&	9.19\\
3&	4.13&	5.06&	5.73&	7.27\\
4&	3.75&	4.03&	7.80&	8.94\\
5&	5.14&	5.96&	5.86&	7.18\\
6&	6.01&	6.95&	8.46&	9.11\\
7&	5.24&	5.62&	8.03&	9.48\\
8&  5.68&	6.93&	10.76&	10.81\\
9&	4.07&	4.41&	5.74&	6.43\\
10&	4.15&	4.15&	5.86&	6.47\\
11&	4.33&	4.77&	7.09&	8.27\\

\hline
Mean&	$4.80 \pm 0.74$ & $5.33 \pm 0.96$  & $7.23 \pm 1.60$ &$8.23 \pm 1.40$\\

    \hline 
  \end{tabular}
  \label{tab:gap} 
\end{table}

\begin{figure}[H]
\centering
\includegraphics[width=0.7\textwidth]{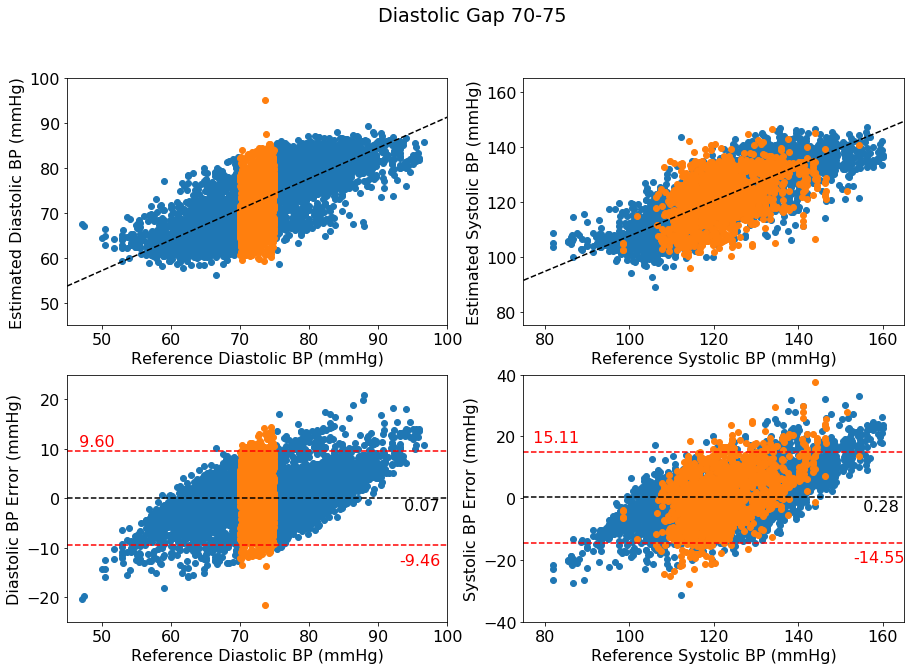}
\caption{Bland-Altman plot for DANN model using four minutes with middle DBP gap}
\label{fig:dbp_mid}
\end{figure}

Figure \ref{fig:dbp_mid} further illustrates this finding, showing pooled test predictions for all users, including for diastolic gaps of 5 mmHg located at 70-75 mmHg. In these plots, the orange points represent samples that fell within the excluded range (in-gap samples) and the blue points represent samples from outside of the gaps. The left side of each figure shows the diastolic pressures, and the right side of each figure shows the systolic pressures.  As can be seen, omitting a range of diastolic pressures does not clearly omit a range of systolic pressure, reflecting the lack of simple relationship between diastolic and systolic pressures.

\begin{figure}[H]
\centering
\includegraphics[width=0.7\textwidth]{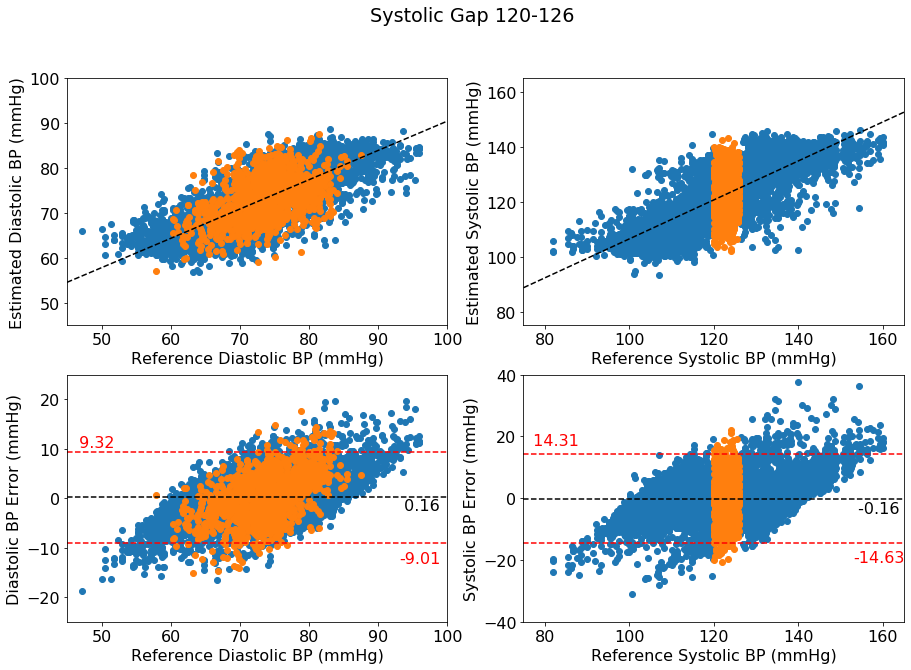}
\caption{Bland-Altman plot for DANN model using four minutes with middle SBP gap}
\label{fig:sbp_mid}
\end{figure}

Similarly, a plot of gaps in systolic blood pressures are shown in Figure \ref{fig:sbp_mid} of 6 mmHg gaps located at 125-131 mmHg.
As before, the orange points represent samples that fell within the excluded range (in-gap samples) and the blue points represent samples from outside of the gaps.  The left side of each figure shows the diastolic pressures, and the right side of each figure shows the systolic pressures.

When a gap falls in the middle of the blood pressure distribution, both low and high blood pressures are equally trained, and when a gap falls at an extreme range, one side is trained well and the side with reduced data is trained poorly. This unbalanced training for an extreme gap results in higher difficulty of generalization. Therefore, the middle gap has fewer errors than gaps in low and high blood pressure ranges and why the in-gap RMSE reported in Table \ref{tab:gap} sees a slight increase compared to the overall RMSE.


To further test the generalizability of this model, we explored the impact of other gap sizes.  We tested diastolic gap sizes of 3 mmHg, 5 mmHg, 7 mmHg, and 10 mmHg and systolic gap sizes of  5 mmHg, 6 mmHg, 7 mmHg, and 10 mmHg. Results of this test on Subject 1 are shown in Table \ref{tab:gap_size}.  As expected,
increasing gap size results in higher RMSE for both overall and in-gap evaluations. In particular, this model struggles in generalizing across gaps of 7 mmHg or larger. For the samples tested here and the data length available, generalizations across this gap size appear to be ill-advised. While this paper centrally focuses on the duration of data needed, aiming to reduce data collection burdens on new users, additional work is needed to identify if further reductions are possible, including the type of blood pressure data needed, such as only low and high blood pressure values. While not an explicit goal of this work, DANN finds preliminary results indicating gaps are possible, and because of the distribution of training data available from other subjects, indicates low and high blood pressure values are more important to provide for the new user than middle (normal) blood pressure values.

\begin{table}[H]
  \footnotesize
  \centering 
  \caption{Generalization results for varying gap sizes applied to Subject 1.  As would be expected, increasing gap size results in poorer performance.}
  \begin{tabular}{cccc}
Gap Type & Gap Size & Overall RMSE & In-Gap RMSE \\ \hline
DBP&    3&  4.58&   4.87\\
DBP&    5&  4.64&   5.18\\
DBP&    7&  4.85&   6.69\\
DBP&    10& 5.59&   6.38\\
\hline
SBP&    5&  5.96&   6.24\\
SBP&    6&  5.99&   6.36\\
SBP&    7&  6.21&   7.07\\
SBP&    10& 6.37&   8.24\\

\hline
  \end{tabular}
  \label{tab:gap_size} 
\end{table}

\section*{A.3 MTL Beat-to-Beat Performance Per Subject with 80\% Training Data}

While the primary focus of this work is on the performance of this model with the application of DANN, we separately studied the performance of the isolated MTL model.
For each subject, we split the data to be 80\% as training set, 10\% as validation set, and 10\% as test set. We test the whole dataset without repetition from 10-fold cross-validation.  This experiment shows overfitting: while the training set is modeled with high average correlation and low average RMSE, the test set suffers significantly in comparison. Performances on the test set are shown in Table \ref{tab:80mtl}. These values still provide a basis for modeling of blood pressure but demonstrate the need for more intelligently trained models, such as DANN in this work.

\begin{table}[H]
  \footnotesize
  \centering 
  \caption{MTL beat-to-beat performance per subject with 80\% training data for diastolic and systolic blood pressure (DBP \& SBP) RMSE (mmHg) and R.}
  \begin{tabular}{ccccc}
    Subject & DBP RMSE & SBP RMSE & DBP R & SBP R \\ \hline
1&	4.40&	5.84&	0.48&	0.29\\
2&	5.40&	8.55&	0.54&	0.63\\
3&	3.95&	5.86&	0.42&	0.58\\
4&	4.14&	7.55&	0.11&	0.33\\
5&	5.29&	5.73&	0.23&	0.27\\
6&	6.15&	8.16&	0.25&	0.49\\
7&	4.94&	7.84&	0.40&	0.48\\
8&	5.30&	10.93&	0.40&	0.50\\
9&	3.70&	5.49&	0.42&	0.22\\
10&	4.06&	5.65&	0.25&	0.23\\
11&	4.30&	7.18&	0.47&	0.54\\

\hline
Mean&	$4.69 \pm 0.73$ & $7.16 \pm 1.68$ & $0.36 \pm 0.13$ & $0.41 \pm 0.15$\\

    \hline 
  \end{tabular}
  \label{tab:80mtl} 
\end{table}

\end{document}